\documentclass{article}

\usepackage{natbib}
     \usepackage[preprint]{nips_2018}

\usepackage[utf8]{inputenc} 
\usepackage[T1]{fontenc}    
\usepackage{hyperref}       
\usepackage{url}            
\usepackage{booktabs}       
\usepackage{amsfonts}       
\usepackage{nicefrac}       
\usepackage{microtype}      
\usepackage{bm}
\usepackage{amsmath}
\usepackage{xcolor}
\usepackage{graphicx}
\usepackage{algorithm}
\usepackage[noend]{algpseudocode}
\usepackage{subfig}

\title{Training CNNs faster with Dynamic Input \\ and Kernel Downsampling}

\author{%
  Zissis Poulos\\
  Department of Electrical and Computer Engineering\\
  University of Toronto\\
  Toronto, CA\\
  \texttt{zpoulos@eecg.toronto.edu} \\
  \And
  Ali Nouri\\
  Department of Electrical and Computer Engineering\\
  University of Toronto\\
  Toronto, CA\\
  \texttt{ali.nouri@mail.utoronto.ca} \\
  \AND
  Andreas Moshovos\\
  Department of Electrical and Computer Engineering\\
  University of Toronto\\
  Toronto, CA\\
  \texttt{moshovos@eecg.toronto.edu} \\
}

\begin{document}

\maketitle

\begin{abstract}
We reduce training time in convolutional networks (CNNs) with a method that, for some of the mini-batches: a) scales down the resolution of input images via downsampling, and b) reduces the forward pass operations via pooling on the convolution filters. Training is performed in an interleaved fashion; some batches undergo the regular forward and backpropagation passes with original network parameters, whereas others undergo a forward pass with pooled filters and downsampled inputs. Since pooling is differentiable, the gradients of the pooled filters propagate to the original network parameters for a standard parameter update. The latter phase requires fewer floating point operations and less storage due to the reduced spatial dimensions in feature maps and filters. 
The key idea is that this phase leads to smaller and approximate updates and thus slower learning, but at significantly reduced cost, followed by passes that use the original network parameters as a refinement stage.
Deciding how often and for which batches the downsmapling occurs can be done either stochastically or deterministically, and can be defined as a training hyperparameter itself. Experiments on residual architectures show that we can achieve up to 23\% reduction in training time with minimal loss in validation accuracy.
\end{abstract}

\section{Introduction}
The abundance of large annotated image datasets has sparked a boom in the applicability of convolutional neural networks (CNNs) on visual tasks, but it also necessitates developing and prototyping large capacity models to capture complex input distributions. This capacity increase comes with much higher demands in terms of training resources. The computational and data footprint cost of state-of-the-art CNN models may exceed billions of floating-point operations (FLOPs) and dozens of gigabytes (GBs), respectively, especially in a mini-batch setting, which is the most common mode of operation when training such models~\citep{resnet1,resnet2,inception,alexnet}.

Distributing the process is one straightforward approach to handle the vast size of the compute and peak memory encountered during training~\citep{cluster}. Training is split into sub-tasks, usually by partitioning input batches. Then, training on these partitions is delegated as separate processes to multiple cluster nodes, be it multiple graphics processing units (GPUs) or custom hardware ({\em e.g}, ASICs). Such approaches however obligate that large multi-core and/or multi-GPU systems are available on demand, which is not always the case. Particularly in prototyping stages, model architectures rapidly change and hyperparameters are tuned to determine what model instance will eventually be deployed. Therefore, evaluating whether training 
converges to acceptable accuracy levels -or if learning even takes place- must happen in short cycles and ideally without the need
to utilize large compute platforms. Moreover, distributed training ideally requires extremely large batch sizes to fully utilize
cluster nodes. However, arbitrarily increasing the batch size may jeopardize or stall convergence, and therefore 
batches are in practice reduced in size, effectively limiting the number of cluster nodes that can be utilized~\citep{Gupta:2017, Krizhevsky14}.

Consequently, a long thread of research is devoted to reducing the cost of training by viewing the problem from an algorithmic or model-level perspective, irrespective of the hardware platform(s) invoked for training. The major challenge
that these methods face is preserving the representational capacity of a target architecture, while reducing training time.
We can broadly categorize these methods based on the property they target towards reducing total training time: 
a) fast convergence methods, and b) per-iteration compute/memory reduction methods. We note that methods between
these two categories can often be combined to attain cumulative benefits.

{\bf Fast convergence:} These methods focus on achieving faster convergence and they effectively reduce the number of iterations (or epochs) required to reach or surpass state-of-the-art accuracy in various datasets. In this context, batch-normalization is a widely adopted method that maintains the stability of feature extraction distributions across layers, allowing deeper architectures with good convergence rates~\citep{batchnorm}. More recently, it was empirically discovered that estimating optimal learning rates on-the-fly and applying those during training instead of using the fixed training policies that are traditionally used, can cause some models to converge extremely fast; a phenomenon coined {\em super-convergence}~\citep{super}. Finally, knowledge transfer has been proposed as a method to convey the parameters
of a small network to a deeper and/or wider model instance via function preserving transformations~\citep{net2net}. This allows learning in the larger
model to commence with initial parameters that would otherwise be learned with much more expensive iterations during the first 
epochs of training.

{\bf Per-iteration compute/memory reduction}: Early techniques in this category reduce the number of FLOPs per iteration, either with more efficient implementations of the convolution operation~\citep{winograd,FFT}, or by introducing low-rank approximations
of the convolution kernels that enable cost-efficient computations~\citep{Jaderberg2014}. Another class of methods performs weight pruning or quantization
during training reducing the number of multiplications in the forward and backward passes, the number of weight updates and
storage requirements as well~\citep{prunetrain, NIPS2016_6372}. More recent methods apply transformations to the target architecture that downsample the feature maps
and/or the convolution filters to reduce execution time early in the training phase and eventually resize those to the dimensions required by the target architecture towards later epochs~\citep{Gusmao}. 

Related techniques have also been applied for reducing the inference cost of deployed models
and are applicable when inference is performed in resource-constrained environments (limited compute and/or memory bandwidth)~\citep{kuen2018stochastic, effnet, Teerapittayanon_2016}.

Despite being successful in reducing CNN training time, several of these methods may require that some desirable properties of the
training process or the target architecture itself are sacrificed. Specifically, 
pruning and quantization methods almost always cause degradation in validation accuracy compared
to the baseline networks~\citep{prunetrain, NIPS2016_6372}. Techniques for reduced-cost implementation of convolution operations maximize their
efficacy when kernels are generally large or the mini-batch size is relatively small, which limits the ability to
explore various architectures and the hyperparameter space~\citep{winograd,FFT}. Knowledge transfer methods, on the other hand, do mot impose 
such limitations, but they do require training multiple smaller instances of a target architecture, constructing a pool of such trained models (or require that 
it exists {\em a priori}), and selecting 
the one that performs best to transplant its weights to the larger model~\citep{net2net}. 
Finally, existing downsampling methods use heuristics to determine the dowsampling factor when resizing convolution kernels
and apply fixed spatial scaling to upsample back to the original shape~\citep{Gusmao}. This process is inherently lossy in terms of
accuracy and spatial resolution and thus constrains the applicability of these transformations; they must be applied only during early stages of training 
so that enough time is available to recuperate from any accuracy loss. In turn, this limits gains regarding total training time, despite
the fact that these transformations can dramatically reduce FLOPs per iteration. 

In this work, we overcome some of the above limitations with a method that: 
a) enables a significant portion of the training iterations to be performed at very low-cost (measured in FLOPs), 
b) can apply these iterations during early and late epochs in the training phase, c) requires no pre-training on a separate
smaller instance of the target architecture, and d) does not degrade validation accuracy.
A low-cost iteration is formed by downsampling the input images of the corresponding mini-batch and all the convolution kernels 
involved in the forward pass. The key novelty behind the proposed formulation is that kernel downsampling is done via
differentiable transformations on the kernels of the target architecture. Differentiable transformations allow gradients 
of the downsampled kernels to propagate backwards to the target architecture's parameters without lossy rescaling. 
Essentially, we learn parameter updates for the target architecture indirectly, using the downsampled kernels as
representative samples for the forward pass computation, requiring much fewer FLOPs for inference and gradient computation.
This permits us to apply these transformations more aggressively at various epochs during training and interleave them with iterations that use the original shape kernels and full resolution images.
Experiments on CIFAR-10 and CIFAR-100 datasets with residual architectures show that we can achieve up to 31\% reduction in training time with no loss in validation accuracy.

\section{Methodology}

There is a twofold rationale behind the proposed approach to reduce training costs. 
First, a downsampled image tends to convey similar input features compared to its full resolution counterpart and it preserves most of the necessary input information for classification tasks~\citep{ChrabaszczLH17}. Pre-training on downsampled images followed by transfer learning to full resolution datasets is common in literature~\citep{net2net, Hinton2015DistillingTK}. Of course, aggressive downsampling may remove some of the distinctive features of the full resolution image ({\em e.g.,} blurred edges, merged image blobs) and cause accuracy deterioration, thus it needs to be done cautiously. Second, for a downsampled image the pixel neighborhood that captures parts of image content useful to classification is proportionally smaller than the neighborhood that captures the same content in the full resolution image. That is, if an input image $x$ with dimensions $H \times W \times D$ is downsampled with a ratio of $r$ (some divisor of $H$ and $W$ other than $1$) to produce a lower resolution image $\hat{x}$ with dimensions $(H/r) \times (W/r) \times D$, then a $K \times K \times D$ input volume in $x$ is represented by a $(K/r) \times (K/r) \times D$ volume in $\hat{x}$. Therefore, we posit that reducing the receptive field of convolution filters also by a factor of (approximately) $r$ per dimension is reasonable and can further reduce the number of operations. The transformation that is applied on filter weights to achieve this reduction is of course of paramount importance towards preserving the intended feature extraction capability of the original filters.

This spatial correlation between downsampled inputs and convolution kernels can be exploited to reduce the computation
during a forward pass, either for a single input image or for a mini-batch. When fewer parameters and activations
are involved in the forward computation, then the compute and storage requirements of the backpropagation algorithm
are also reduced. In what follows we describe a method to perform such low-cost training iterations. When the network performs a low-cost iteration we refer to it as {\em low-mode} operation. On the other hand, when the kernel and input shapes are unmodified 
we say that the network is in {\em full-mode} operation. 

\subsection{Full-mode Operation}
In full-mode, the network performs all computations that are dictated by its specification. 
We restrict our attention to convolutional layers, since these usually dominate execution time
and are the ones that this method modifies dynamically during the training process.

Given an input image $\bm{I}$ let function $\bm{y}=f(\bm{I};\bm{\theta)}$ represent the network in full-mode operation, where
$\bm{I}$ is the input image, $\bm{y}$ is the output of the network which is a distribution of conditional probabilities 
over class categories given $\bm{I}$, and $\bm{\theta}$ is the collection of the network's learnable parameters (or weights). 
Also, let $\bm{h}^{(i)} = \phi(\bm{W^{(i)}}\bm{h}^{(i-1)})$ represent convolutional layer $i$ in full-mode, where
$\bm{h}^{(i-1)}$ is an input tensor of activations to the layer, $\bm{W}^{(i)}$ is the weight matrix, $\phi$ is an
activation function ({\em e.g.,} ReLU), and $\bm{h}^{(i)}$ is the output tensor of activations. 

To derive an asymptotic cost for  convolutional layer $i$, let the dimensions of the input activation tensor be such that $\bm{h}^{(i-1)} \in \mathbb{R}^{H^{(i-1)} \times W^{(i-1)} \times C^{(i-1)}}$, where $H^{(i-1)}, W^{(i-1)}$ and $C^{(i-1)}$ denote the height, width and channel depth, respectively. If the first layer of the network ($i=1$) is a convolutional layer then we have $\bm{h}^{(0)} \equiv \bm{I}$, with $\bm{I} \in \mathbb{R}^{H \times W \times C}$. Finally, the weight matrix $\bm{W}^{(i)}$ can be represented as a collection of $N^{(i)}$ kernels (or filters) of size $K^{(i)} \times K^{(i)} \times C^{(i-1)}$, each of which is convolved with $\bm{h}^{(i-1)}$ to produce a single output activation in $\bm{h}^{(i)}$. The asymptotic complexity of layer $i$ can then be written as $\mathcal{O}\big{(} (K^{(i)})^2 C^{(i-1)} H^{(i)} W^{(i)} C^{(i)}\big{)}$.

In practice, it is common to train a CNN in a mini-batch setting, where forward computation is performed in batches of 
input images. In each training iteration $t$, the network accepts a batch of input images denoted $\bm{B}_{t} = 
\{\bm{I}_1, \dots, \bm{I}_{|\bm{B}_t|}\}$. Updating the network's parameters after iteration $t$ is related to the following minimization:

\begin{align}
& \min_{\bm{\theta}_{t}} \bm{L}(\bm{\theta}_{t}) \\ \nonumber
\bm{L}(\bm{\theta}_{t}) & = \frac{1}{|\bm{B}_t|}\sum\limits_{j=1}^{|\bm{B_t}|} \bm{L}(\bm{t}_j, f(\bm{I_j};\bm{\theta}_{t})) 
\end{align}

\noindent where $\bm{L}(\bm{t}_j, f(\bm{I_j};\bm{\theta}_{t}))$ is the loss function quantifying the error between the ground truth label
$\bm{t}_j$ of input $\bm{I}_j$ and the network's output 
$f(\bm{I_j};\bm{\theta_t)}$ when the network's parameters in iteration $t$ are $\bm{\theta}_t$. 
The backpropagation algorithm is then applied to compute gradient updates for all network parameters~\citep{backprop}.
Here, we focus again on the update step for convolutional layer $i$. Suppose that $\bm{W^{(i)}}_{t-1}$ is the weight matrix of layer $i$ 
at the end of iteration $t-1$ (beginning of iteration $t$). Then, the weight matrix is updated to $\bm{W^{(i)}}_{t}$ using the steepest
gradient descent method, as follows:

\begin{equation}
\bm{W}^{(i)}_{t} \leftarrow \bm{W}^{(i)}_{t-1} - \epsilon \frac{\partial \bm{L}(\bm{W}^{(i)}_{t-1})}{\partial \bm{W}^{(i)}_{t-1}} 
\end{equation}

\noindent where $\bm{L}(\bm{\theta}_{t-1})$ is the cost to minimize over mini-batch $\bm{B}_t$, and $\epsilon$ is 
the learning rate. 

\subsection{Low-mode Operation}
In low-mode operation we force the network to perform a forward pass at a drastically decreased cost by:
a) reducing the spatial dimensions of the input images and b) reducing the spatial dimensions of the kernels in 
each convolutional layer $i$. We note that the number of channels $C^{(i-1)}$ of the input activations and the number of filters $N^{(i)}$
remain unchanged in this mode.

\subsubsection{Input Downsampling}
We use $\hat{\bm{I}} \overset{r}{\leftarrow} \bm{I}$ to denote that image $\bm{I} \in \mathbb{R}^{H \times W \times C}$ is downsampled to image $\hat{\bm{I}} \in \mathbb{R}^{(H/r) \times (W/r) \times C}$ with a ratio $r$ per dimension, where $r$ is
some divisor of $H$ and $W$ other than $1$. This spatial transformation can be performed with various techniques, including Gaussian blurring followed by row/column elimination, or bi-linear interpolation~\citep{Nixon:2008}. We give preference to the former since it behaves well for $r>2$. The notation implictly applies when we refer to batches of images, and in that case we use $\hat{\bm{B}_t} \overset{r}{\leftarrow} \bm{B}_t$ to say that all images in batch $\bm{B}_t = \{ \bm{I}_1, \dots, \bm{I}_{|\bm{B}_t|} \}$ are downsampled by a ratio of $r$ to create batch $\hat{\bm{B}}_t=\{ \hat{\bm{I}}_1, \dots, \hat{\bm{I}}_{|\hat{\bm{B}}_t|} \}$.

\subsubsection{Kernel Downsampling} 
Given a kernel $\bm{F}^{(i)}$ in layer $i$ with size $K^{(i)} \times K^{(i)} \times C^{(i-1)}$ we create kernel $\hat{\bm{F}}^{(i)}$ with dimensions
$\lceil K^{(i)}/r \rceil \times \lceil K^{(i)}/r \rceil \times C^{(i-1)}$ using a {\em differentiable} spatial transformation $T(\cdot)$, so that $T(\bm{F}) = \hat{\bm{F}}$. We have experimented with various transformations, observing different effects in accuracy.
The transformation we propose is an average pooling operator that produces filters with odd dimensions.
For example, if $r=2$ and $\bm{F}$ is a $5 \times 5 \times C$ filter then $\hat{\bm{F}} = T(\bm{F}) = AvgPool(\bm{F})$, where $AvgPool(\cdot)$ is 
a $2 \times 2$ pooling operator with stride $2$ and asymmetric padding. The resulting filter $\hat{\bm{F}}$ has dimensions $3\times 3 \times C$. Alternatively we can generate $\lfloor K/r \rfloor \times \lfloor K/r \rfloor \times C$ filters, depending on the architecture. Indicatively, we use $3 \times 3$ average pooling to transform $3 \times 3$ convolutions to $1 \times 1$ pointwise convolutions.

Downsampling both input and convolution kernels may cause a cascading effect of spatially reduced feature maps
in intermediate layers. The ratio of this (indirect) downsampling for input activations can be controlled through the
stride parameter. Suppose that in low-mode, the input activations of layer $i$ are 
$\hat{\bm{h}}^{(i-1)} \in \mathbb{R}^{H^{(i-1)}/r^{(i-1)} \times W^{(i-1)}/r^{(i-1)} \times C^{(i-1)}}$, 
where $r^{(i-1)}$ is the reduction factor compared to the spatial dimensions of the full-mode activation inputs
$\bm{h}^{(i-1)}$. An analogous reduction can be obtained for the output activations as well, so that these become
$\hat{\bm{h}}^{(i)} \in \mathbb{R}^{H^{(i)}/r^{(i)} \times W^{(i)}/r^{(i)} \times C^{(i)}}$.

The asymptotic complexity of layer $i$ in low-mode can then be expressed as $\mathcal{O}\big{(} (K^{(i)}/r)^2 C^{(i-1)} (H^{(i)}/r^{(i)}) (W^{(i)}/r^{i}) C^{(i)}\big{)}$. In a relative sense, the low-mode operation for layer $i$ involves
$\mathcal{O}\big{(} (r)^2 (r^{(i)})^2 \big{)}$ less computation compared to the full-mode operation.

{
\begin{figure}[t!]\centering
\resizebox{4.2in}{!}{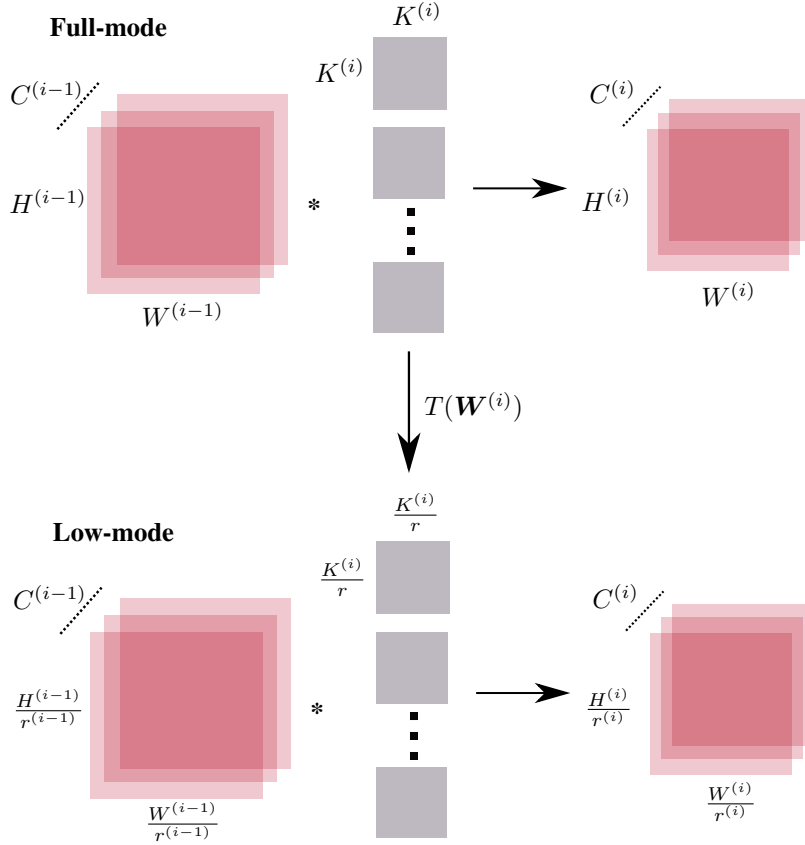}
\caption{Full-mode vs. Low-mode forward computation for a convolutional layer} 
\label{fig:conv} 
\end{figure}
}

\section{Learning in Low-mode}
With these transformations the network now computes some function $\hat{\bm{y}}=\hat{f}(\hat{\bm{I}};\hat{\bm{\theta)}}$, where
$\hat{\bm{y}} \neq \bm{y}$, in general, and $\hat{\bm{\theta}}$ is the new set of (transformed) parameters.
For clarity we can use the transformation $T(\cdot)$ on $\bm{\theta}$ to mean that the relevant parameters (convolution weights)
are transformed to generate $\hat{\bm{\theta}}$. That is, we write $\hat{\bm{\theta}}=T(\bm{\theta})$.
Similarly, we say that $\hat{\bm{W}}^{(i)} = T({\bm{W}}^{(i)})$ is the weight matrix of the transformed parameters for layer $i$,
{\em i.e.} the collection of all $N^{(i)}$ downsampled kernels of layer $i$ after applying the transformation $T(\cdot)$ with 
downsampling ratio $r$.
The goal then at iteration $t$, if it is performed in low-mode, is to minimize:

\begin{align}
& \min_{\bm{\theta}_t} \bm{L}(T(\bm{\theta}_t)) \\ \nonumber
\bm{L}(T(\bm{\theta}_t)) & = \frac{1}{|\hat{\bm{B}}_t|}\sum\limits_{j=1}^{|\hat{\bm{B_t}}|} \bm{L}(\bm{t}_j, \hat{f}(\hat{\bm{I}}_j;T(\bm{\theta}_t))) 
\end{align}

If iteration $t$ occurs in low-mode, let $\hat{\bm{W}}^{(i)}_{t-1}$ be the weight matrix of the transformed parameters for layer $i$ at the beginning of iteration $t$. Then, layer $i$ computes some non-linear function $\hat{\bm{h}}^{(i)} = \phi(\hat{\bm{W}}^{(i)}_{t-1}\hat{\bm{h}}^{(i-1)})$.
The gradients of $\hat{\bm{W}}^{(i)}_{t-1}$ with respect to the loss in Eq.3, $\partial \bm{L}(\hat{\bm{W}}^{(i)}_{t-1}) / \partial \hat{\bm{W}}^{(i)}_{t-1}$, are computed by backpropagation as usual. The end goal, however, is to derive updates for
weight matrix $\bm{W}^{(i)}$; that is, the network parameters before the application of $T(\cdot)$, so that these can
be used during subsequent full-mode operations. If $T(\cdot)$ is selected such that it is differentiable everywhere ({\em e.g.}, average pooling as previously discussed), then 
$\partial\hat{\bm{W}}^{(i)}_{t-1} / \partial\bm{W}^{(i)}_{t-1} = \partial T(\bm{W}^{(i)}_{t-1}) / \partial\bm{W}^{(i)}_{t-1}$ 
exists and can be used to compute updates for $\bm{W}^{(i)}_{t}$ by an additional step of the backpropagation algorithm:

\begin{equation}
\bm{W}^{(i)}_{t} \leftarrow \bm{W}^{(i)}_{t-1} - \epsilon \frac{\partial \bm{L}(\hat{\bm{W}}^{(i)}_{t-1})}{\partial \hat{\bm{W}}^{(i)}_{t-1}} \frac{\partial \hat{\bm{W}}^{(i)}_{t-1}}{\partial \bm{W}^{(i)}_{t-1}}  
\end{equation}

{
\begin{figure}[t!]\centering
\resizebox{4.2in}{!}{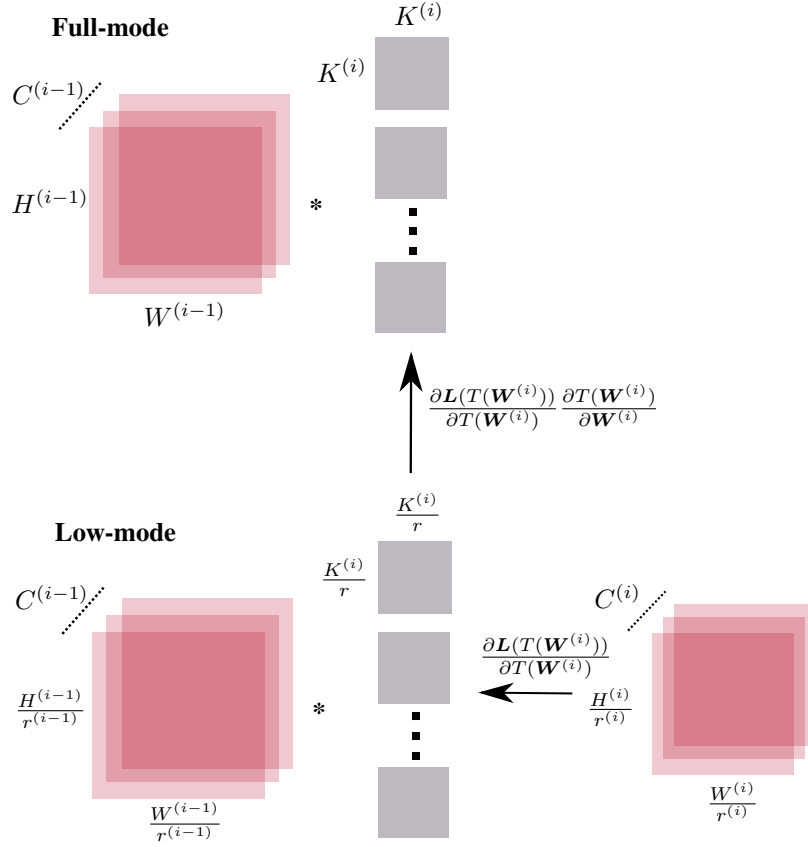}
\caption{Low-mode backward propagation for convolutional layer weights $\bm{W}^{(i)}$} 
\label{fig:conv} 
\end{figure}
}

\subsection{Selecting Low-mode Iterations}

Choosing which iterations and how many are done in low-mode is expected to affect the
convergence rate of the target architecture. Aggressive policies where many contiguous iterations are on
downsampled inputs and kernels may cause severe deterioration in accuracy, from which the model may fail
to recover. In contrast, overly pessimistic policies where low-mode iterations are infrequent may produce
negligible runtime benefits.

In this work, we adopt a stochastic scheme where the model alternates between the two modes with some pre-specified probability, 
determined before training commences. Specifically, we define $p_{low}^{e}$ to be the probability that the next iteration is done
in low-mode within epoch $e$ (the probability may differ across epochs). For instance, assuming that training is divided into $E$ epochs, then setting $p_{low}^{e} =  0.5, 1 \leq e \leq E$ forces the model into low-mode half of the time {\em in expectation}.
However, we found that letting the target model operate
in full-mode for the first few epochs after a learning rate adjustment takes place helps with fast convergence and accuracy refinement,
respectively. Details are further discussed in Section 4. The policy can also be deterministic ({\em e.g.,} alternating based on
a fixed frequency), but we give preference to stochastic schemes to remove any bias that a deterministic policy may introduce.

\subsection{Overall Training Algorithm}

The pseudocode for training a target model $M$ using the method presented here (with downsampling ratio $r$), is given below.
We assume that the per-epoch low-mode probabilities are determined beforehand and gathered into a vector $P$ of length $E$.

\begin{algorithm}
\caption{Training with Downsampling}\label{Algo}
\begin{algorithmic}[1]
\State $\mbox{{\bf Input:} model $M$, downsampling ratio $r$, per-epoch downsampling propabilities $P$}$
\For {$e$ = 1 : $E$}
\State $p_{low}^e \leftarrow P[e]$
\For {$t$ = 1 : $T$}
\State $low \leftarrow Bernoulli(p_{low}^e)$ 
\If {$low == 0$}
\State $\mbox{train model $M$ with batch $\bm{B}_t$, kernels $\bm{F}^{(i)}$ for all convolution layers $i$}$ 
\State $\mbox{update $\bm{\theta}_{t}$}$
\EndIf
\If {$low == 1$}
\State $\hat{\bm{I}}_k \overset{r}{\leftarrow} \bm{I}_k , \forall \bm{I}_k \in \bm{B}_t$, 
\State $\hat{\bm{B}_t} \overset{r}{\leftarrow} \bm{B}_t$
\State $\mbox{$\hat{\bm{W}}^{(i)}_{t-1} = T(\bm{W}^{(i)}_{t-1})$ with ratio $r$ for all convolution layers $i$}$
\State $\mbox{train model $M$ with batch $\hat{\bm{B}}_t$, parameters $\hat{\bm{W}}^{(i)}_{t-1}$ for all convolution layers $i$}$
\State $\mbox{update $\bm{W}^{(i)}_{t}$ using Eq.4 for all convolution layers $i$}$
\EndIf 
\EndFor
\EndFor
\end{algorithmic}
\end{algorithm}

In the process outlined above, the generation of downsampled images (Line 10) can also be done off-line for the entire
training set as long as one ensures batches $B_t$ and $\hat{B}_t$ match after each shuffling round between epochs. The transformation
can be applied on the host side (CPU) in the case where GPUs are used for training, thus it does not have to consume processing time
on the device side. The weight transformations (Line 12) are performed on the device side during the forward pass. Finally, in the pseudocode we suppress the update of weights for fully connected (FC) layers (it is explicit in Line 8 and implied in Line 14), since FC parameters are shared between the two modes of operation.

\section{Experiments}
We evaluate the proposed method on the residual architectures ResNet18 and ResNet50~\citep{resnet1,resnet2} using the CIFAR10 dataset~\citep{cifar10}. We measure runtime on an NVIDIA 1080Ti GPU. A summary of the experimental setting is given below:

\begin{enumerate}
\item Downsampling ratio $r = 2$ ({\em i.e.}, input images have dimensions 16x16x3 in low-mode).
\item Total number of epochs $E=200$, and batch size $|B_t| = |\hat{B}_t| = 125$.
\item Full-mode / Low-mode: we let the first 4 epochs immediately after a learning rate adjustment to run on full-mode exclusively
( $p_{low}^{e} = 0$).
In the residual models studied, learning rate begins at $0.1$ and is adjusted to $0.01$ at epoch 80 and to $0.001$ at epoch 120~\citep{resnet1,resnet2}.
\item The downsampling probability is set to $p_{low}^{e} = p = 0.5$ for all other epochs. 
\item Due to the fact that feature maps have reduced spatial dimensions through all layers in low-mode, we drop the max pooling layer immediately after the first convolutional layer in both ResNet18 and ResNet50 (only in low-mode). 
\item ResNet18: the 3$\times$3 kernels in layers conv1 and conv2 of the basic block cell are transformed to 1$\times$1 kernels as per Section 3.
\item ResNet50: only the 3$\times$3 kernels in layer conv2 of the bottleneck cell are transformed to 1$\times$1 kernels. 
\item FC layers remain unchanged in both models.
\item Other hyper-parameters remain unchanged, including kernel strides and optimizer parameters, such as momentum and weight decay~\citep{resnet1,resnet2}.
\end{enumerate}

\begin{figure}[!t]
\centering
\subfloat[ResNet18 on CIFAR10 ($r=2$)]{\includegraphics[width=0.85\textwidth]{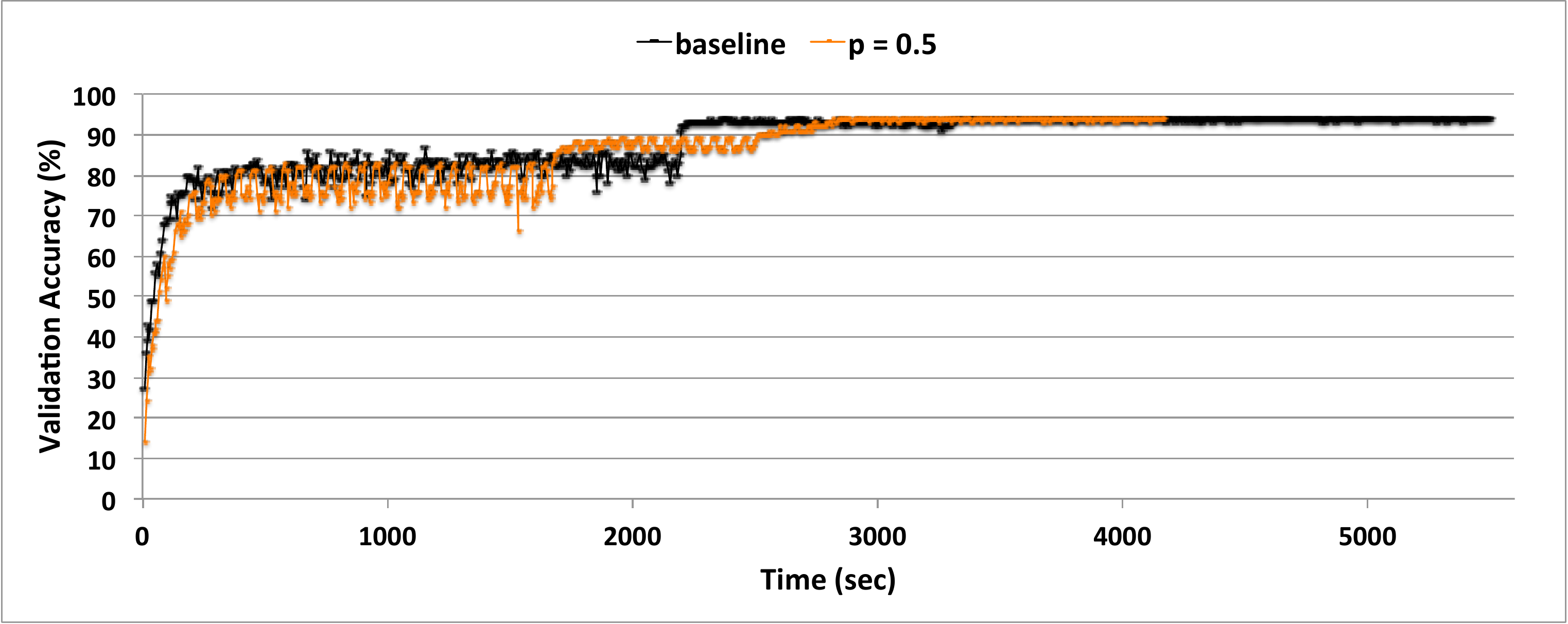}}\,
\subfloat[ResNet50 on CIFAR10 ($r=2$)]{\includegraphics[width=0.85\textwidth]{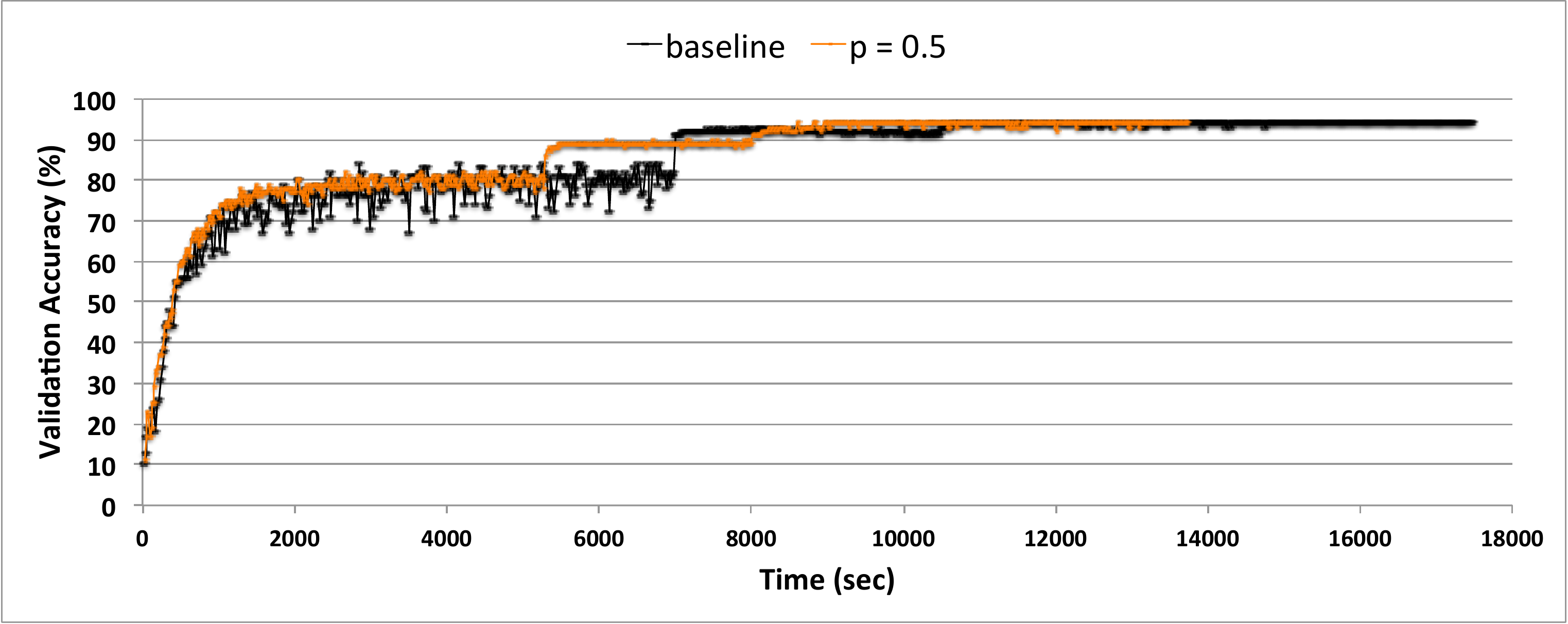}}
\caption{Validation accuracy vs. time, between baseline training and the proposed method}
\label{fig:resnets}
\end{figure}

Figure~\ref{fig:resnets} illustrates the validation accuracy curves achieved on the two models studied when the proposed downsampling
method is applied using the setting above. We compare those to the validation curves of the standard training process, referred to as {\em baseline} in the figure. For the proposed method and the baseline we apply the same training hyper-parameters (summarized above), and in fact ensure that they match the suggested ones in the original publications of these models~\citep{resnet1,resnet2}. The figures plot accuracy over time, which includes GPU time accounting for both forward and backward passes and host time for transforming the input images.

\begin{figure}[!t]
\centering
\subfloat[ResNet18 on CIFAR10 ($r=2$)]{\includegraphics[width=0.85\textwidth]{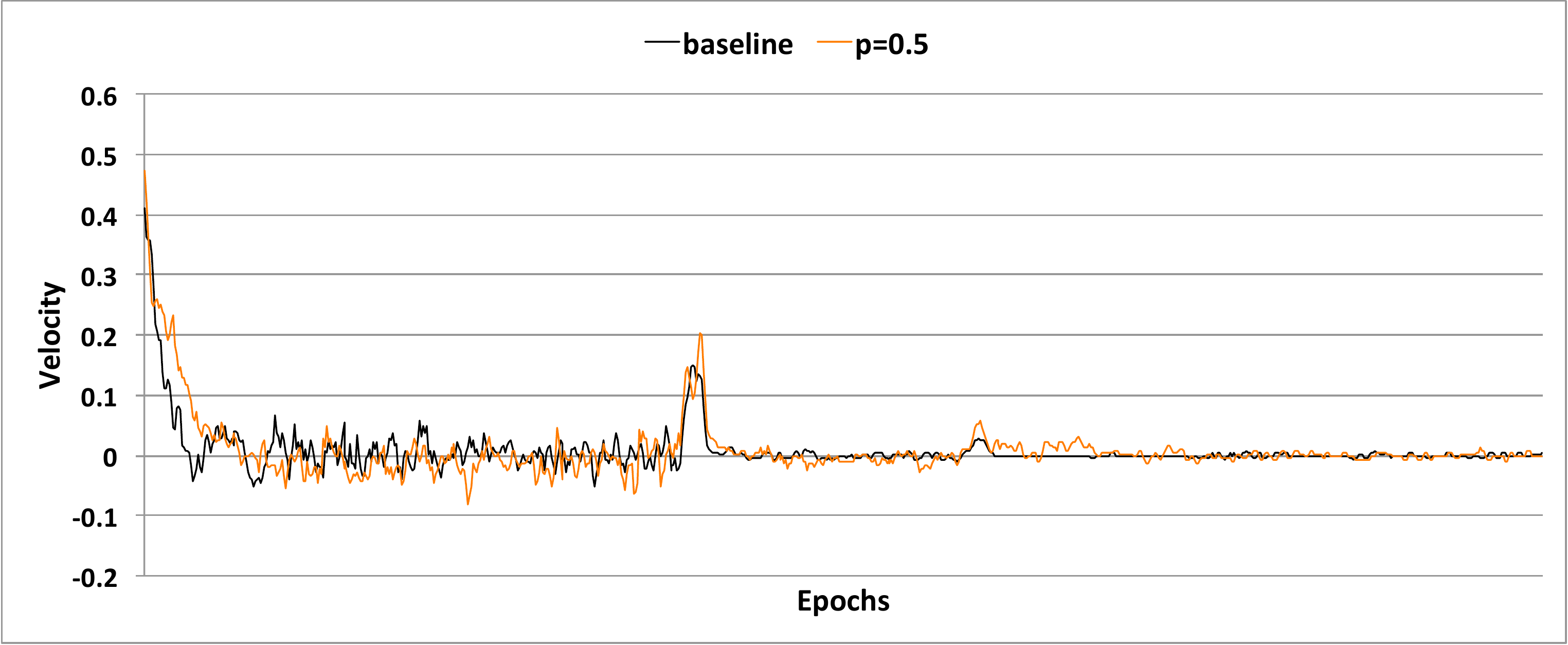}}\,
\subfloat[ResNet50 on CIFAR10 ($r=2$)]{\includegraphics[width=0.85\textwidth]{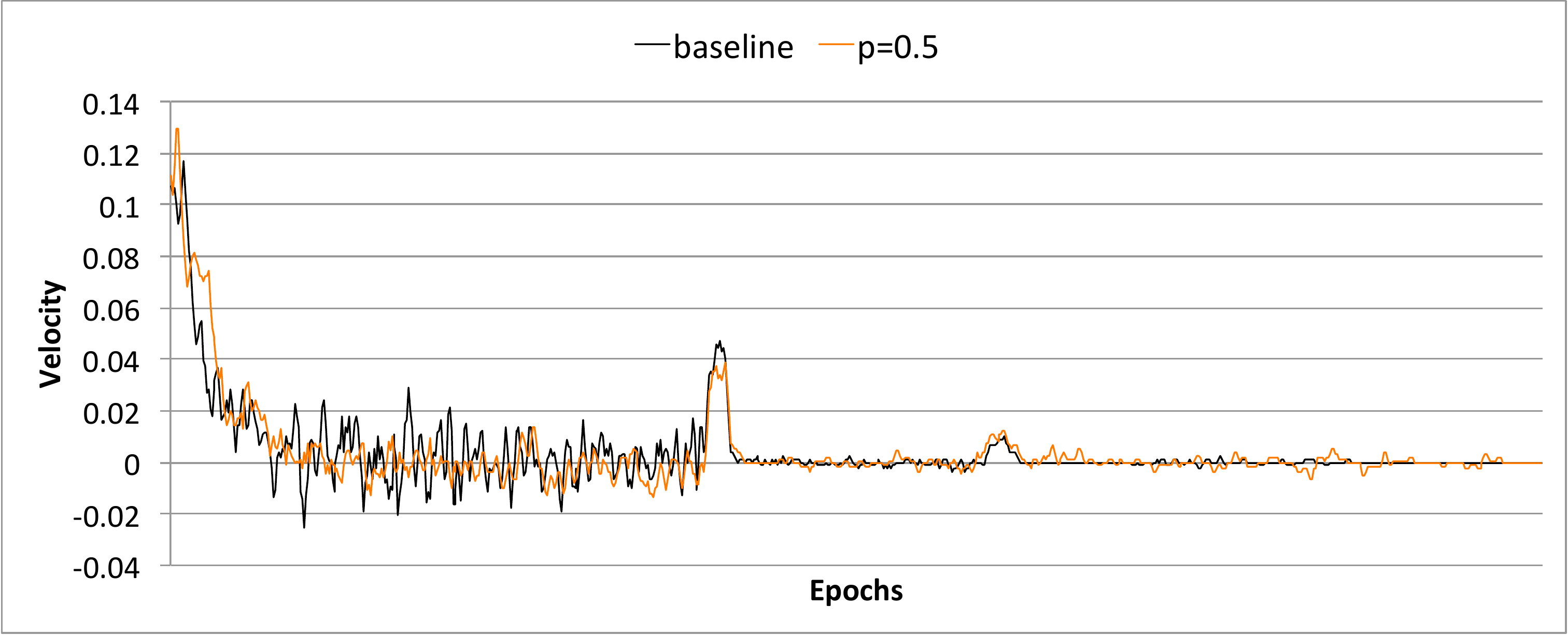}}
\caption{Velocity per epoch, between baseline training and the proposed method}
\label{fig:resnetsV}
\end{figure}

The extent of the proposed method's success can be based on two criteria: a) whether it converges to the same validation accuracy 
as the baseline, and b) whether the validation curve is time-shifted to the left, {\em i.e.} convergence to the final (maximum) accuracy is achieved earlier in the proposed method. As Figure~\ref{fig:resnets}(a) shows, the proposed method on ResNet18 (with downsamping probability $p=0.5$) converges to 94\% accuracy at around 2,800 seconds ($\approx$ 47 minutes), while the baseline method hits the same
accuracy levels at 3,400 seconds ($\approx$ 57 minutes). This constitutes an 18\% reduction in runtime. For a deeper and larger capacity model however, such as ResNet50, runtime gains are more pronounced. As shown in Figure~\ref{fig:resnets}(b), downsampling on ResNet50
with $p=0.5$ leads to the same accuracy as the baseline, but does so 23\% faster ($\approx 150$ vs. $\approx 193$ minutes). 
Faster convergence in ResNet50 vs. ResNet18 (in a relative sense) cannot be merely attributed to compute reduction per-layer in low-mode operation; these are similar between the two models. Indicatively, a single iteration during low-mode consumes 39\% of a full-mode iteration in ResNet50 and 37\% in ResNet18. Thus, this behavior could instead rise from the robustness of the deeper architecture: the deeper architecture is expected to be less sensitive to the weight transformations in low-mode (less likely to
observe a drop in accuracy) and can also recover better from any loss if it occurs. 

To illustrate this phenomenon, Figure~\ref{fig:resnetsV} demonstrates how learning progresses for ResNet18 and ResNet50 when the proposed
downsampling method is applied.
Specifically, learning progress is quantified by measuring the velocity of training and how this changes throughout epochs.
Here, velocity is the quotient $\frac{\Delta A}{\Delta\tau} = \frac{A' - A}{\tau' - \tau}$, where $A', A$ are validation accuracies
at times $\tau', \tau$. In other words, we measure the change in accuracy over fixed periods of time. The figure reports the moving
average of velocity in windows of 3 epochs (1200 iterations). In the first epoch, downsampling has a greater negative effect in validation accuracy for both models. In turn, full mode operations are able to recover the model's accuracy fairly quickly. The oscillations of $\frac{\Delta A}{\Delta\tau}$ in Figure~\ref{fig:resnetsV} capture this phenomenon. For early training epochs (10-80) where this phenomenon is more pronounced, oscillations in ResNet50
range between $[-0.01,0.01]$, whereas in ResNet18 they range between $[-0.08,0.03]$ and are lopsided towards negative effects. This
could imply that ResNet50 is less sensitive to the transformations performed during low-mode.

Finally, Table~\ref{tb:res} summarizes the accuracy and training time reduction achieved with the proposed method on the 
two models studied. Note that the final relative error introduced by the proposed method is below 1\%. As previously discussed, time savings are
larger for the deeper architecture, despite the fact that per-iteration savings are similar between the two models.

\begin{table}
\centering
\caption {Training time reduction on CIFAR10 (p=0.5)}
\begin{tabular}{c||c|c|c|c}
\hline\noalign{\smallskip}
        \bf{Model} & error (baseline) & error (proposed) & total time reduction & iteration time reduction  \\ \hline \hline
        ResNet-18 & 6.7\% & 6.9\% & 18\% & 63\% \\ \hline
        ResNet-50 & 6.3\% & 6.4\% & 23\% & 61\%\\ \hline
    \end{tabular}
    \label{tb:res}
\end{table}

\section{Conclusions and Future Work}
We present a method for reducing training time in CNNs with minimal impact in model accuracy.
The method achieves these train-time savings by downsampling input images and kernels for a significant portion of
the total number of training iterations. The method permits learning to progress even in this low-cost mode by propagating
gradients from the downsampled kernels to the kernels of the target architecture that are then used to recover accuracy.
For a wider application of this method, it is important to evaluate the extent to which it generalizes well for other architectures and different data regimes (e.g., Imagenet) and tasks (e.g., segmentation, detection).

\bibliographystyle{unsrtnat}
\bibliography{nips_2018}

\end{document}